\pdfoutput=1

\documentclass[11pt]{article}

\usepackage[final]{acl}

\usepackage{times}
\usepackage{latexsym}

\usepackage[T1]{fontenc}

\usepackage[utf8]{inputenc}

\usepackage{microtype}
\usepackage{booktabs}
\usepackage{soul}
\sethlcolor{red}
\usepackage{multirow}
\usepackage{CJKutf8}

\usepackage{inconsolata}

\usepackage{graphicx}

\title{CCNU at SemEval-2025 Task 3: \\Leveraging Internal and External Knowledge of Large Language Models for Multilingual Hallucination Annotation}

\author{Xu Liu {\normalfont and} Guanyi Chen\thanks{Corresponding Author} \\
  Hubei Provincial Key Laboratory of Artificial Intelligence and Smart Learning, \\
  National Language Resources Monitoring and Research Center for Network Media, \\
  School of Computer Science, Central China Normal University \\
  \texttt{liuxu@mails.ccnu.edu.cn, g.chen@ccnu.edu.cn}}

\begin{document}
\maketitle
\begin{abstract}
We present the system developed by the Central China Normal University (CCNU) team for the Mu-SHROOM shared task, which focuses on identifying hallucinations in question-answering systems across 14 different languages. Our approach leverages multiple Large Language Models (LLMs) with distinct areas of expertise, employing them in parallel to annotate hallucinations, effectively simulating a crowdsourcing annotation process. Furthermore, each LLM-based annotator integrates both internal and external knowledge related to the input during the annotation process. Using the open-source LLM DeepSeek-V3, our system achieves the top ranking (\#1) for Hindi data and secures a Top-5 position in seven other languages. In this paper, we also discuss unsuccessful approaches explored during our development process and share key insights gained from participating in this shared task.
\end{abstract}

\section{Introduction}

Hallucinations refer to content in outputs that neither follow from the inputs nor are supported by known facts. In 2024, \citet{mickus-etal-2024-semeval} organized a shared task on detecting hallucinations in machine translation, definition modelling, and paraphrasing systems. Building on this foundation and expanding to a new domain—question answering—SemEval-2025 Task 3~\citep[Mu-SHROOM;][]{vazquez-etal-2025-mu-shroom} broadens the scope of hallucination detection. This task extends beyond English to cover 14 different languages and moves beyond binary classification (i.e., determining whether an item contains hallucinations) to pinpointing the exact location of hallucinations, as illustrated in Table~\ref{tab:example}.

Although Large Language Models (LLMs) inevitably produce hallucinations~\citep{xu2024hallucination}, they have also proven effective in detecting them: four of the six highest-scoring systems in the 2024 challenge leveraged state-of-the-art LLMs~\citep{mickus-etal-2024-semeval}. However, the new task setting introduced above presents two key challenges for these LLM-based solutions. 

\textbf{First}, Mu-SHROOM shifts the focus from hallucinations in generation systems, such as machine translation and paraphrasing, to hallucinations in question-answering (QA) systems. This shift alters the definition of hallucination. As discussed in~\citet{thomson2020gold,duvsek2020evaluating,ji2023survey,deemter-2024-pitfalls}, hallucinations in generation systems refer to outputs that contradict the given inputs. In contrast, within QA, hallucinations pertain to outputs that contradict corresponding ``facts''. Consequently, detecting hallucinations in a given QA pair requires a model first to determine what constitutes the relevant ``facts''. Since these facts are not explicitly present in the input, the model must be capable of integrating knowledge from multiple sources.

\begin{table}[t]
    \centering
    \small
    \begin{tabular}{cp{4.5cm}}
    \toprule
    \textbf{Question} & What did Petra van Staveren win a gold medal for? \\
    \midrule
    \textbf{Answer} & Petra van Stoveren won a \hl{silver} medal in the \hl{2008} Summer Olympics in \hl{Beijing, China}.\\
    \bottomrule
    \end{tabular}
    \caption{An example test item from Mu-SHROOM. The hallucinations are coloured in red.}
    \label{tab:example}
\end{table}

\begin{figure*}[t!]
    \centering
    \includegraphics[scale=0.24]{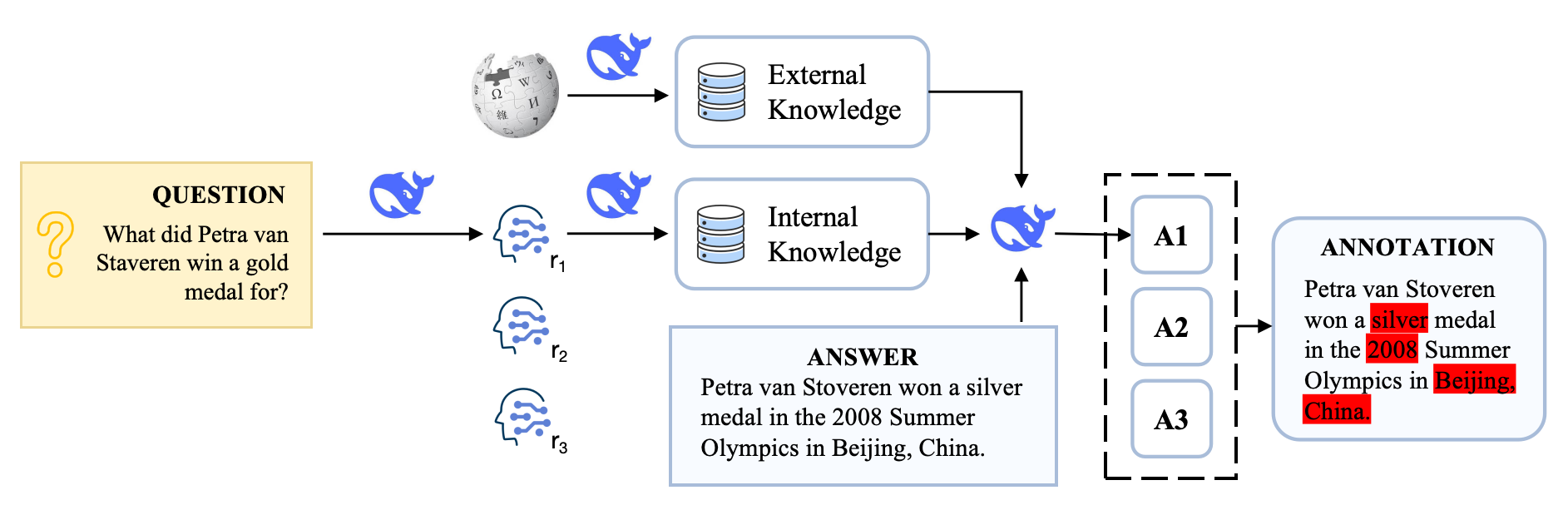}
    \caption{An overview of our hallucination annotation system. The blue whale represents LLM (i.e., DeepSeek).}
    \label{fig:overview}
\end{figure*}

\textbf{Second}, the fine-grained hallucination annotation scheme in Mu-SHROOM increases the likelihood of annotation disagreements. Different annotators may label the same error in different ways. For example, consider the error ``silver'' in Table~\ref{tab:example}: the term is incorrect because Petra van Stroveren won a gold medal in the Olympic Games. However, one annotator might highlight only the word ``silver'', while another might annotate the entire noun phrase ``a silver medal''. Such disagreements are natural, and Mu-SHROOM addresses them by employing multiple annotators and resolving inconsistencies through majority voting. This collaborative approach is difficult to replicate with a single LLM-based hallucination detector.

Following the approach of the 2024 challenge winners, our solution employs LLMs with optimizations to address the two challenges discussed above. To tackle the first issue, our LLM-based hallucination detector retrieves relevant ``facts'' not only from its internal knowledge but also from external resources, thereby integrating both internal and external knowledge. To address the second issue, our solution mimics the crowdsourced annotation process by leveraging multiple LLMs, assigning them different roles, and having them annotate each QA pair in parallel before reaching a consensus through voting. Notably, our approach requires no fine-tuning or language-specific optimizations. Using an open-source LLM—DeepSeek-V3~\citep{liu2024deepseek}—as the backbone, our solution achieved \#1 ranking on Hindi data and placed in the Top 5 for Arabic, Basque, Catalan, Czech, English, Persian, and Spanish. \footnote{For German and French, we used GPT-4o, ranking \#3 and \#15, respectively.}

\section{The Mu-SHROOM Task}

The Mu-SHROOM task~\citep{vazquez-etal-2025-mu-shroom} asks systems to annotate hallucinations in QA in 14 languages, including Arabic, Basque, Catalan, Chinese, Czech, English, Farsi, Finnish, French, German, Hindi, Italian, Spanish, and Swedish. The annotations contain: (1) \textbf{Hard Labels}, i.e., hallucinations in QA pairs as in Table~\ref{tab:example}; and (2) \textbf{Soft Labels}, i.e., probability of each token in the answer being a hallucination term. 

Mu-SHROOM evaluates each system using Intersection-over-Union (IoU) for hard labels and Spearman correlation (Cor) for soft labels. See \citet{vazquez-etal-2025-mu-shroom} for more details.

\section{Methodology}

This section starts with explaining how we prompt LLMs to annotate hallucinations in QA systems, followed by how we make them leverage internal and external knowledge during annotation. Figure~\ref{fig:overview} provides an overview of our hallucination annotation system. 

\subsection{Prompting LLMs to Mark Hallucinations} \label{sec:prompt}

\begin{figure*}[t!]
    \centering
    \fbox{\includegraphics[scale=0.26]{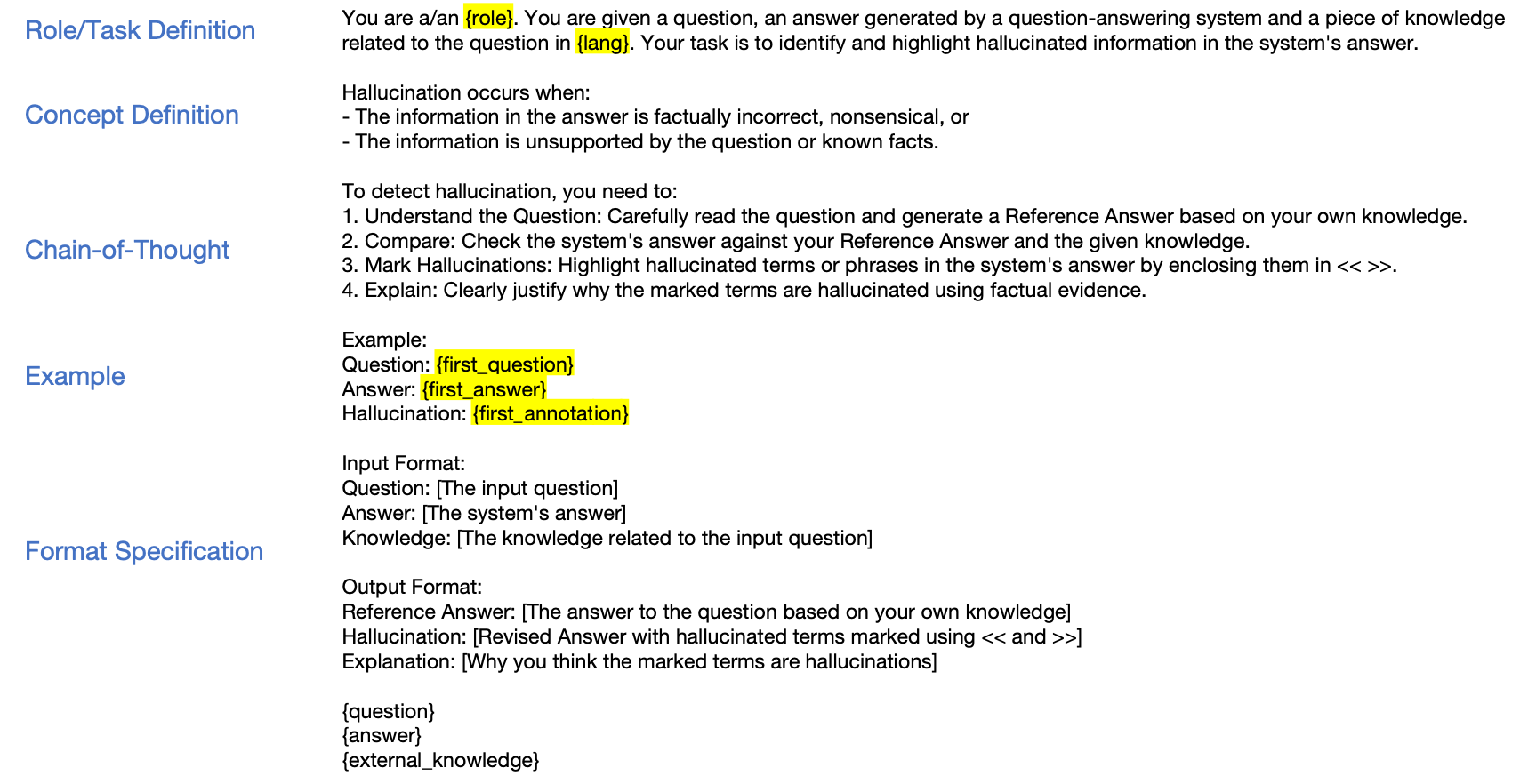}}
    \caption{The main prompt in our system. The variables highlighted in yellow will be replaced with their corresponding desired values.}
    \label{fig:prompt}
\end{figure*}

As shown in Figure~\ref{fig:prompt}, our prompt\footnote{For all languages, the prompt is always in English, with the only modification being the replacement of `{lang}' with the name of the test language.} begins by defining the task and the concept of hallucination to provide the LLM with a clearer understanding of the background. Notably, we refine the definition of hallucination in the context of QA by specifying that hallucinated content is characterized as ``factually incorrect'', ``nonsensical'', or ``not supported by known facts''.

We then incorporate a Chain-of-Thought (CoT), outlining the steps the LLM should follow to improve hallucination annotation. In this CoT, we first instruct the LLM to generate a reference answer based on the given question (see further discussion in Section~\ref{sec:internal_k}). It then compares the provided answer with the generated reference answer to identify hallucinated content. We also ask the LLM to explain why the annotated terms are classified as hallucinations.

Next, we present the LLM with an example, extracted from the first item in the development set. We also specify the expected input and output format. It is worth mentioning that rather than directly returning soft and hard labels—where hallucinations are represented as integer indices indicating their start and end positions—we instruct the LLM to mark hallucinated terms within the answer. This is achieved by having the LLM generate a revised version of the answer, where hallucinated terms are enclosed within $\langle \langle$' and $\rangle \rangle$'. This approach bypasses the LLM's limited ability to accurately count indices. Finally, we provide the LLM with the input QA pair along with external knowledge (see further discussion in Section~\ref{sec:external_k}).

For each QA pair, we prompt the LLM 12 times and obtain 12 annotations. For each token in a given answer, we calculate the probability of it being hallucinated by computing the proportion of times it was annotated as a hallucination across the 12 annotations.

\subsection{Internal Knowledge} \label{sec:internal_k}

We compel the LLM to leverage its internal knowledge when processing a given question by first requiring it to generate an answer based solely on its own knowledge and then annotate hallucinations accordingly. To further diversify the internal knowledge used in this process, we assign the LLM different roles across the 12 runs. This is achieved by employing another LLM to determine a set of distinct roles (i.e., $r_i$ in Figure~\ref{fig:overview}), each capable of evaluating the factual accuracy of the given QA pair and detecting potential hallucinations. Additionally, we instruct this role-assigning LLM to ensure that the suggested roles are as diverse as possible. The corresponding prompt for this role assignment process can be found in Appendix~\ref{sec:appendix_prompt}.

\subsection{External Knowledge} \label{sec:external_k}

We extract external knowledge from Wikipedia based on the given question. Specifically, for each QA pair, we first prompt the LLM to identify key terms from the question (the corresponding prompt can be found in Appendix~\ref{sec:appendix_prompt}). These key terms are then used to construct a query for retrieving relevant knowledge from Wikipedia, with the first returned result serving as the external knowledge. Since the retrieved content may be excessively long, we employ another LLM to summarize and refine it, producing the final external knowledge (the prompt for this summarization process is also provided in Appendix~\ref{sec:appendix_prompt}).

\section{Experiments}

\begin{figure}
    \centering
    \includegraphics[width=\linewidth]{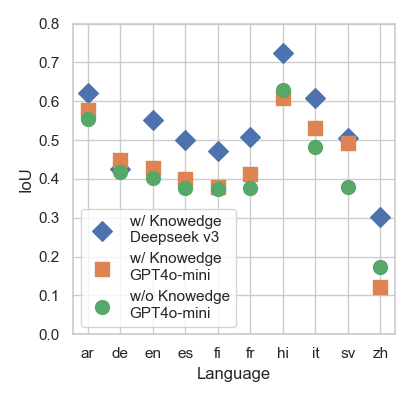}
    \caption{Performance in terms of IoU on 10 languages whose development sets are available. }
    \label{fig:result}
\end{figure}

Figure~\ref{fig:result} presents the performance of our system in terms of IoU, which is considered more important than Cor, across data in 10 languages with available development sets. The figure compares the results of our system using GPT-4o-mini as the backbone LLM, both with and without (internal and external) knowledge, as well as a version employing DeepSeek-V3 with knowledge.

As the results indicate, incorporating both internal and external knowledge consistently improves the LLMs' ability to annotate hallucinations across all 10 languages, with the exception of Chinese. This anomaly is mitigated by replacing GPT-4o-mini with DeepSeek-V3, suggesting that the fundamental capability of the backbone LLM plays a crucial role in extracting high-quality knowledge.

Moreover, we observe that: (1) When comparing DeepSeek-3V to GPT-4o-mini, DeepSeek-3V outperforms GPT-4o-mini in all languages except German; and (2) Our system achieves the highest performance on Hindi data and the lowest performance on Chinese data. Its performance remains relatively consistent across other languages, regardless of whether the language is high-resourced or low-resourced.

\begin{table}[t]
    \centering
    \small
    \begin{tabular}{llcc}
    \toprule
    \textbf{Lang.} & \textbf{Model} & \textbf{IoU} & \textbf{Cor} \\
    \midrule
    \multirow{3}{*}{EN} & DeepSeek-V3 & \textbf{55.04} & 55.88 \\
    & GPT-4o-mini & 42.74 & 53.27 \\
    & GPT-4o & 52.04 & \textbf{63.27} \\ \midrule
    \multirow{3}{*}{FR} & DeepSeek-V3 & 50.70 & \textbf{51.68} \\
    & GPT-4o-mini & 41.37 & 47.57 \\
    & GPT-4o & \textbf{57.75} & 50.55 \\ \midrule
    \multirow{3}{*}{ZH} & DeepSeek-V3 & \textbf{30.11} & 27.02 \\
    & GPT-4o-mini & 12.17 & \textbf{27.87} \\
    & GPT-4o & 22.30 & 26.78 \\
    \bottomrule
    \end{tabular}
    \caption{Performance of our system for English, French, and Chinese with different backbone LLMs.}
    \label{tab:llm}
\end{table}

\paragraph{The Choice of Backbone LLMs.} As mentioned earlier, the choice of backbone LLM is crucial for effectively leveraging knowledge. To further investigate this, we conducted a small experiment on English, French, and Chinese data, comparing three backbone LLMs: DeepSeek-V3, GPT-4o-mini, and GPT-4o. Surprisingly, the open-sourced DeepSeek-V3 not only performs best for Chinese (which is expected, given that a Chinese company developed it) but also outperforms the other models for English. The results in Figure~\ref{fig:result} further highlight its strong performance for low-resourced languages.

The final decision relies on both the performance and the cost. In our case, an experiment on data in a single language costs \$10 using GPT-4o but merely \$0.15 using DeepSeek-V3. As a result, we finally used GPT-4o for German and French (see results in Figure~\ref{fig:result} and Table~\ref{tab:llm}) and DeepSeek-V3 for all other languages. 

\begin{table}[t]
    \centering
    \small
    \begin{tabular}{lcclcc}
    \toprule
    \textbf{Lang.} & \textbf{IoU} & \textbf{Rank} & \textbf{Lang.} & \textbf{IoU} & \textbf{Rank} \\
    \midrule
    Arabic & 59.95 & 5/29 & Catalan & 66.94 & 2/21 \\
    Czech & 48.52 & 5/23 & German & 59.17 & 3/28 \\
    English & 53.94 & 5/41 & Spanish & 51.25 & 4/32 \\
    Basque & 57.85 & 3/23 & Persian & 66.00 & 4/23 \\
    Finnish & 51.17 & 13/27 & French & 48.23 & 15/30 \\
    Hindi & 74.66 & 1/24 & Italian & 70.60 & 7/28 \\
    Swedish & 50.45 & 15/27 & Chinese & 38.34 & 18/26 \\
    \bottomrule
    \end{tabular}
    \caption{Performance of our system on the test sets in terms of IoU and rank.}
    \label{tab:test}
\end{table}

\paragraph{Results on the Test Sets.} Table~\ref{tab:test} reports the performance of our system on the test sets. It achieved \#1 ranking on Hindi data and placed in the Top 5 for the other 8 languages. Consistent with the results on the development sets, the system showed the lowest performance on the Chinese test set (see Section 6 for a potential explanation).

\paragraph{The Effect of Marking Hallucinations in Place.} As mentioned in Section~\ref{sec:prompt}, our system asks LLMs to mark hallucinations directly in the given QA pairs instead of returning the indices of the starting and ending positions of hallucinations. An experiment using Llama-3.1-8B reveals that this improves IoU from 33.68 to 39.97 on English data. 

\section{Unsuccessful Approaches}

In this section, we discuss the unsuccessful approaches encountered during the development of our system.

\paragraph{Ignoring Typos.} Through analysing the annotations generated by LLMs, we found that they often classify typos and grammatical errors as hallucinations, and such errors are rarely treated as hallucinations in the corpus. To address this, we instructed the LLMs to ignore typos and grammatical mistakes. However, in an experiment using Llama-3.1-8B, this adjustment led to a decrease in IoU from 39.97 to 29.12 on English data. This decline suggests that LLMs may struggle to differentiate between typos and hallucinations, as both are perceived as forms of error.

\paragraph{Correcting before Annotating.} Our system leverages internal knowledge by prompting the LLM to generate an answer to the input question based on its own knowledge before annotating hallucinations. We experimented with an alternative strategy: instructing the same LLM in two separate runs. In the first run, the LLM was asked to only generate an answer from its own knowledge. This generated response was then used in the second run to assist in annotating hallucinations. While this approach achieved a similar IoU score to our final solution, it was more computationally expensive due to the additional LLM invocation. Therefore, we ultimately abandoned this strategy.

\paragraph{Incorporating External Knowledge without Summarising.} Our system incorporates external knowledge by first extracting relevant information from Wikipedia and then summarizing it using an LLM. The summarization step was introduced to mitigate potential issues arising from overly lengthy or irrelevant extracted content with respect to the QA pairs awaiting annotation. Considering the inherent trade-off between information volume and density, where summarization increases information density but reduces overall content, we tested the removal of the LLM-based summarization step. However, an experiment using Qwen2.5-14B revealed that eliminating summarization decreased the IoU score from 42.55 to 38.24 on the English dataset.

\section{Discussion}

\paragraph{Quality of the Dataset.}

Our system performs surprisingly poorly on Chinese data (see Figure~\ref{fig:result}). Interestingly, other participants in this shared task seem to face a similar issue, as the baseline approach—which indiscriminately marks all terms as hallucinations—ranks 7th out of 26 teams~\citep{vazquez-etal-2025-mu-shroom}. Upon examining the Chinese dataset, we identified problematic cases, with the following serving as an example:
\begin{CJK*}{UTF8}{gbsn}
\begin{quote}
安德列·克拉克夫（Andrei Konchalovsky）是一位俄罗斯导演、编剧和\textcolor{red}{制片}人，他的作品包括：《\textcolor{red}{俄罗斯方舟》（2011年}）、《\textcolor{red}{悲悯世界》(1991}) 、 《\textcolor{red}{莫斯科不相信眼泪》 (18}\% \textcolor{red}{白人) (Moskva slezam ne verit, 1\% blondynki})（\textcolor{red}{10\%的白人}）等。
\end{quote}
\end{CJK*}
This is a problematic data instance because: (1) It exhibits degeneration~\citep{holtzmancurious}, making it difficult for annotators to determine which parts should be labelled as hallucinations; and (2) It contains numerous inconsistencies. For example, symbols like `\%' and `)' are sometimes marked as hallucinations, while in other cases, they are not.

\paragraph{Comparing the Results on Hard and Soft Labels.} 
We compute the Mean Reciprocal Rank (MRR) of our systems on the final rankings in terms of both IoU and Cor, and obtain 0.26 and 0.34, respectively. This means that our system has better performance in deciding soft labels than hard labels. This is probably attributed to our design of letting multiple LLMs mimic the crowdsourcing annotation process.  

\paragraph{Definition of Hallucination.}

According to~\citet{vazquez-etal-2025-mu-shroom}, the definition of hallucination given to the annotators is:
\begin{quote}
\textbf{Hallucination}: content that contains or describes facts that are not supported by the provided reference. In other words, hallucinations are cases where the answer text is more specific than it should be, given the information available in the provided context.
\end{quote}
For us, this definition poses several issues: (1) The second half of the definition leans more towards describing over-specification rather than hallucination. Its reasoning aligns closely with the Gricean Maxim of Quantity~\citep{grice1975logic} rather than the Maxim of Quality, as discussed in~\citet{deemter-2024-pitfalls}. This discrepancy also creates an inconsistency between the two parts of the definition. (2) This Gricean-style definition (i.e., ``more specific than it should be'') is inherently vague, as the appropriate level of specificity is subjective and uncertain for annotators (see \citet{chen2023varieties} for discussions). For example, in Table~\ref{tab:example}, one could argue that specifying ``Beijing, China'' is redundant, as ``2007 Summer Olympics'' already serves as an unambiguous referring expression.

\section{Conclusion}

This paper presents the Central China Normal University (CCNU) team's solution to SemEval-2025 Task 3, the Mu-SHROOM task, which requires submissions to annotate hallucinations in question-answering systems across 14 different languages. Our approach employs multiple LLMs with distinct roles, prompts them in parallel to annotate hallucinations in order to simulate a crowdsourcing annotation process. Each LLM-based annotator integrates both internal and external knowledge related to the input during the annotation process. A small ablation study highlights the importance of incorporating knowledge. Finally, we report several unsuccessful attempts and share key observations gained from participating in this shared task.

In future, we plan to have a closer look at how the choice of different roles would influence the performance of our system and seek an annotation scheme that handles disagreements better (see Section~6) and considers severities of different kinds of hallucinations~\citep{van-miltenburg-etal-2020-gradations}.

\bibliography{custom}

\begin{thebibliography}{12}
\providecommand{\natexlab}[1]{#1}

\bibitem[{Chen and van Deemter(2023)}]{chen2023varieties}
Guanyi Chen and Kees van Deemter. 2023.
\newblock Varieties of specification: Redefining over-and under-specification.
\newblock \emph{Journal of Pragmatics}, 216:21--42.

\bibitem[{Du{\v{s}}ek and Kasner(2020)}]{duvsek2020evaluating}
Ond{\v{r}}ej Du{\v{s}}ek and Zden{\v{e}}k Kasner. 2020.
\newblock Evaluating semantic accuracy of data-to-text generation with natural
  language inference.
\newblock In \emph{Proceedings of the 13th International Conference on Natural
  Language Generation}, pages 131--137.

\bibitem[{Grice(1975)}]{grice1975logic}
Herbert~P Grice. 1975.
\newblock Logic and conversation.
\newblock In \emph{Speech acts}, pages 41--58. Brill.

\bibitem[{Holtzman et~al.()Holtzman, Buys, Du, Forbes, and
  Choi}]{holtzmancurious}
Ari Holtzman, Jan Buys, Li~Du, Maxwell Forbes, and Yejin Choi.
\newblock The curious case of neural text degeneration.
\newblock In \emph{International Conference on Learning Representations}.

\bibitem[{Ji et~al.(2023)Ji, Lee, Frieske, Yu, Su, Xu, Ishii, Bang, Madotto,
  and Fung}]{ji2023survey}
Ziwei Ji, Nayeon Lee, Rita Frieske, Tiezheng Yu, Dan Su, Yan Xu, Etsuko Ishii,
  Ye~Jin Bang, Andrea Madotto, and Pascale Fung. 2023.
\newblock Survey of hallucination in natural language generation.
\newblock \emph{ACM computing surveys}, 55(12):1--38.

\bibitem[{Liu et~al.(2024)Liu, Feng, Xue, Wang, Wu, Lu, Zhao, Deng, Zhang, Ruan
  et~al.}]{liu2024deepseek}
Aixin Liu, Bei Feng, Bing Xue, Bingxuan Wang, Bochao Wu, Chengda Lu, Chenggang
  Zhao, Chengqi Deng, Chenyu Zhang, Chong Ruan, et~al. 2024.
\newblock Deepseek-v3 technical report.
\newblock \emph{arXiv preprint arXiv:2412.19437}.

\bibitem[{Mickus et~al.(2024)Mickus, Zosa, Vazquez, Vahtola, Tiedemann,
  Segonne, Raganato, and Apidianaki}]{mickus-etal-2024-semeval}
Timothee Mickus, Elaine Zosa, Raul Vazquez, Teemu Vahtola, J{\"o}rg Tiedemann,
  Vincent Segonne, Alessandro Raganato, and Marianna Apidianaki. 2024.
\newblock \href {https://doi.org/10.18653/v1/2024.semeval-1.273}
  {{S}em{E}val-2024 task 6: {SHROOM}, a shared-task on hallucinations and
  related observable overgeneration mistakes}.
\newblock In \emph{Proceedings of the 18th International Workshop on Semantic
  Evaluation (SemEval-2024)}, pages 1979--1993, Mexico City, Mexico.
  Association for Computational Linguistics.

\bibitem[{Thomson and Reiter(2020)}]{thomson2020gold}
Craig Thomson and Ehud Reiter. 2020.
\newblock A gold standard methodology for evaluating accuracy in data-to-text
  systems.
\newblock In \emph{Proceedings of the 13th International Conference on Natural
  Language Generation}, pages 158--168.

\bibitem[{van Deemter(2024)}]{deemter-2024-pitfalls}
Kees van Deemter. 2024.
\newblock \href {https://doi.org/10.1162/coli_a_00509} {The pitfalls of
  defining hallucination}.
\newblock \emph{Computational Linguistics}, 50(2):807--816.

\bibitem[{van Miltenburg et~al.(2020)van Miltenburg, Lu, Krahmer, Gatt, Chen,
  Li, and van Deemter}]{van-miltenburg-etal-2020-gradations}
Emiel van Miltenburg, Wei-Ting Lu, Emiel Krahmer, Albert Gatt, Guanyi Chen, Lin
  Li, and Kees van Deemter. 2020.
\newblock \href {https://doi.org/10.18653/v1/2020.inlg-1.45} {Gradations of
  error severity in automatic image descriptions}.
\newblock In \emph{Proceedings of the 13th International Conference on Natural
  Language Generation}, pages 398--411, Dublin, Ireland. Association for
  Computational Linguistics.

\bibitem[{V\'azquez et~al.(2025)V\'azquez, Mickus, Zosa, Vahtola, Tiedemann,
  Sinha, Segonne, S\'anchez-Vega, Raganato, Libovický, Karlgren, Ji, Helcl,
  Guillou, de~Gibert, Bengoetxea, Attieh, and
  Apidianaki}]{vazquez-etal-2025-mu-shroom}
Ra\'ul V\'azquez, Timothee Mickus, Elaine Zosa, Teemu Vahtola, J\"org
  Tiedemann, Aman Sinha, Vincent Segonne, Fernando S\'anchez-Vega, Alessandro
  Raganato, Jindřich Libovický, Jussi Karlgren, Shaoxiong Ji, Jindřich
  Helcl, Liane Guillou, Ona de~Gibert, Jaione Bengoetxea, Joseph Attieh, and
  Marianna Apidianaki. 2025.
\newblock \href {https://helsinki-nlp.github.io/shroom/} {Sem{E}val-2025 {T}ask
  3: {Mu-SHROOM}, the multilingual shared-task on hallucinations and related
  observable overgeneration mistakes}.

\bibitem[{Xu et~al.(2024)Xu, Jain, and Kankanhalli}]{xu2024hallucination}
Ziwei Xu, Sanjay Jain, and Mohan Kankanhalli. 2024.
\newblock Hallucination is inevitable: An innate limitation of large language
  models.
\newblock \emph{arXiv preprint arXiv:2401.11817}.

\end{thebibliography}

\newpage
\appendix

\section{Further Prompts} \label{sec:appendix_prompt}

\begin{table}[htbp]
    \centering
    \small
    \caption{Prompt for assigning roles during internal knowledge extraction.}
    \begin{tabular}{p{0.95\linewidth}}
        \toprule
The task is when given a pair of a question and an answer in \{lang\}, to try to identify up to 5 distinct expert identities capable of evaluating the factual accuracy of the answer and detecting potential hallucinations. Ensure the suggested identities are diverse and tailored to the specific context of the input.\\[5px]

Given question: \{question\}\\
Given answer: \{answer\}\\[5px]

Please give your output in JSON format with keys `Identities' and `Reason'. \\
Under the content of `Identities', please output the identity, your identity should be correct, clear, and easy to understand.\\
Under the content of `Reason', explain why you output these identities.\\[5px]
Provide a clear and concise response, just give your answer in JSON format as I request, and don’t say any other words.\\
        \bottomrule
    \end{tabular}
\end{table}

\begin{table}[htbp]
    \centering
    \small
    \caption{Prompt for extracting key terms from the input question.}
    \begin{tabular}{p{0.95\linewidth}}
        \toprule
You are given a question and you need to extract a keyword, which will be used for querying Wikipedia.\\[5px]

Input Format:\\
Question: [The input question]\\[5px]

Output Format:\\
Keyword: [A keyword directly extracted from the input question, only the essential terms, usually the name and main topic.]\\[5px]

Example:\\
Question: What did Petra van Staveren win a gold medal for?\\
Keyword: Petra van Staveren\\
        \bottomrule
    \end{tabular}
\end{table}

\begin{table}[htbp]
    \centering
    \small
    \caption{Prompt for summarising and refining the extracted external knowledge.} 
    \begin{tabular}{p{0.95\linewidth}}
        \toprule
You are given a question, an answer, and a set of knowledge in JSON retrieved from Wikipedia in {lang}. We are building a system that detects hallucinations in the given answer. Your task is to refine the given knowledge from Wikipedia to make it helpful to serve as a reference for identifying hallucinations in the answer.\\[5px]

To refine the knowledge, you need to:\\
Analyze the Question: Carefully analyze the question and the answer to identify what is being asked and determine the key information needed to identify the factual errors in the answer.\\
Evaluate the Given Knowledge: Review the related knowledge provided and simultaneously assess its relevance to the question, determining whether it is directly useful, partially useful, or not applicable to identify the factual errors in the answer.\\
Generate Knowledge: Based on the judgment, either refine the provided knowledge, integrate it with new insights, or create a standalone response in EN that contains knowledge that helps identify the fact errors in the answer effectively.\\[5px]

Input:\\
Question: {question}\\
Answer: {answer}\\
Related knowledge: {knowledge}\\[5px]

Please give your output in JSON format with keys `Knowledge' and `Reason'. \\
Under the content of `Knowledge', please output the refined knowledge in a single paragraph.\\
Under the content of `Reason', explain why you make such refinements.\\[5px]
Provide a clear and concise response, just give your answer in JSON format as I request, and don’t say any other words.\\
        \bottomrule
    \end{tabular}
\end{table}

\end{document}